\documentclass{article}
\pdfoutput=1

\usepackage[final]{neurips_2025}
\usepackage{float}

\usepackage[utf8]{inputenc}
\usepackage[T1]{fontenc}
\usepackage{hyperref}
\usepackage{url}
\usepackage{amsfonts}
\usepackage{nicefrac}
\usepackage{microtype}
\usepackage{xcolor}
\usepackage{makecell}
\usepackage{multirow}
\usepackage{listings}
\usepackage{mdframed}
\usepackage{adjustbox}
\usepackage{booktabs}
\usepackage{enumitem}
\usepackage{array}
\usepackage{placeins}
\usepackage{graphicx}

\usepackage{caption}
\usepackage[nameinlink]{cleveref}

\newcommand{\black}[1]{\textcolor{black}{#1}}
\newcommand{\blue}[1]{\textcolor{blue}{#1}}

\newcommand{\blacklink}[2]{\href{#1}{\textcolor{black}{#2}}}

\definecolor{questioncolor}{RGB}{0,0,0}
\definecolor{choicecolor}{RGB}{0,0,255}
\definecolor{contextcolor}{RGB}{255,140,0}

\hypersetup{
  colorlinks=true,
  linkcolor=black,
  citecolor=black,
  urlcolor=blue,
  pdfborder={0 0 0},
  pdftitle={Reasoning Models Achieve 96\% on CFA Exams},
  pdfauthor={Jaisal Patel, Yunzhe Chen, Kaiwen He, Keyi Wang, Kairong Xiao, Xiao-Yang Liu}
}

\newcommand{\openaismall}{\raisebox{-0.25\height}{\includegraphics[height=1.2em]{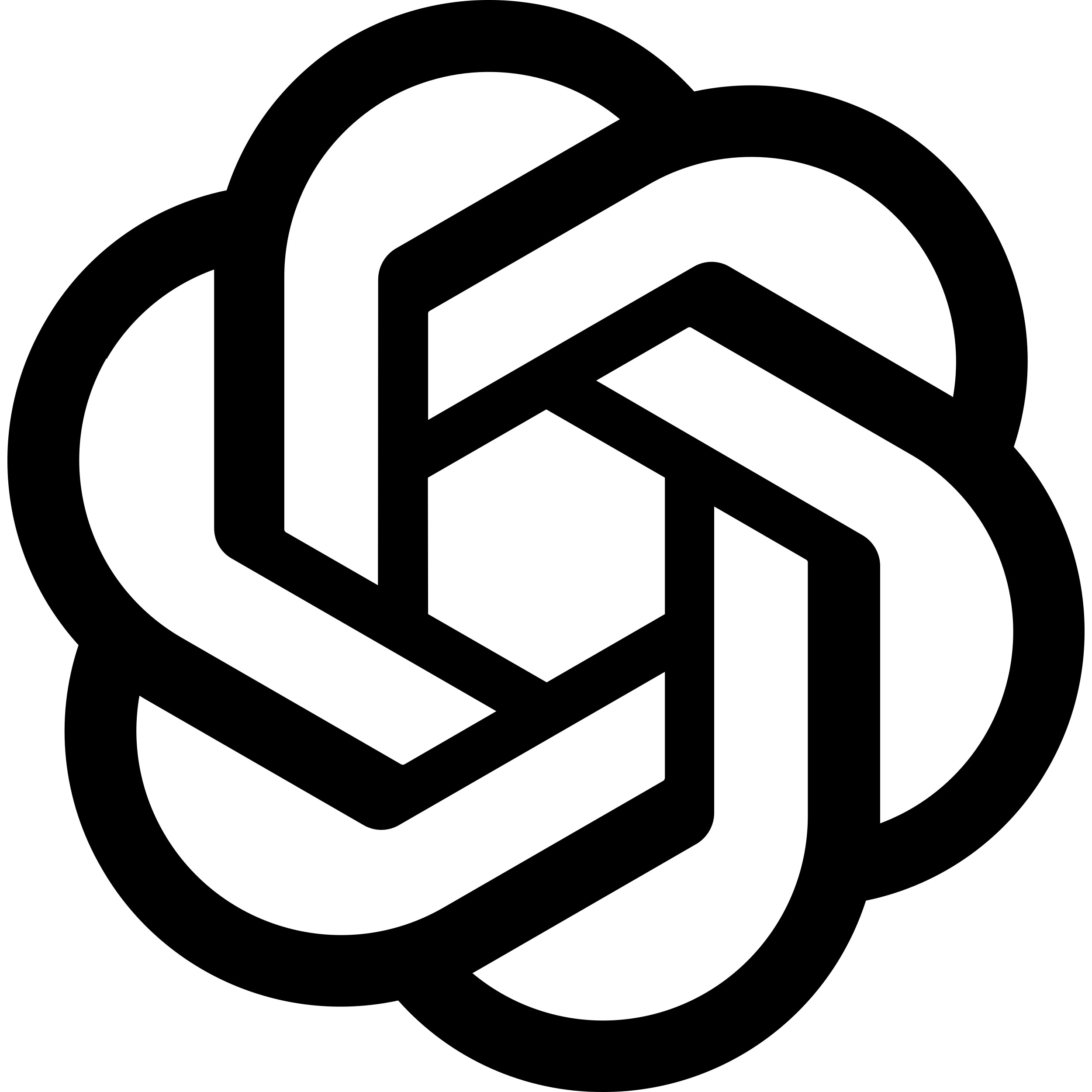}}\hspace{0.2em}OpenAI}
\newcommand{\deepseeksmall}{\raisebox{-0.25\height}{\includegraphics[height=1.2em]{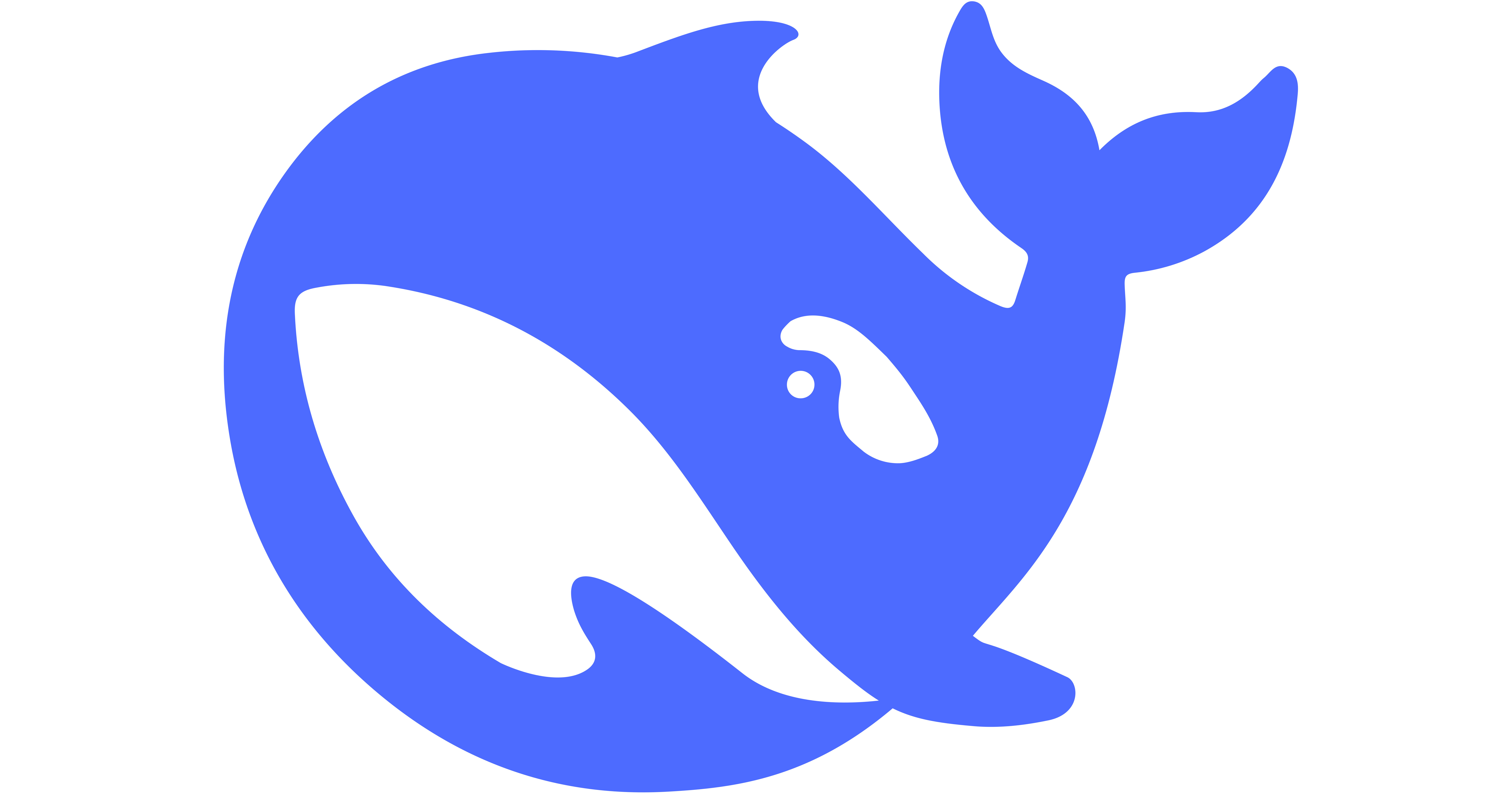}}\hspace{0.2em}DeepSeek}
\newcommand{\xaismall}{\raisebox{-0.25\height}{\includegraphics[height=1.2em]{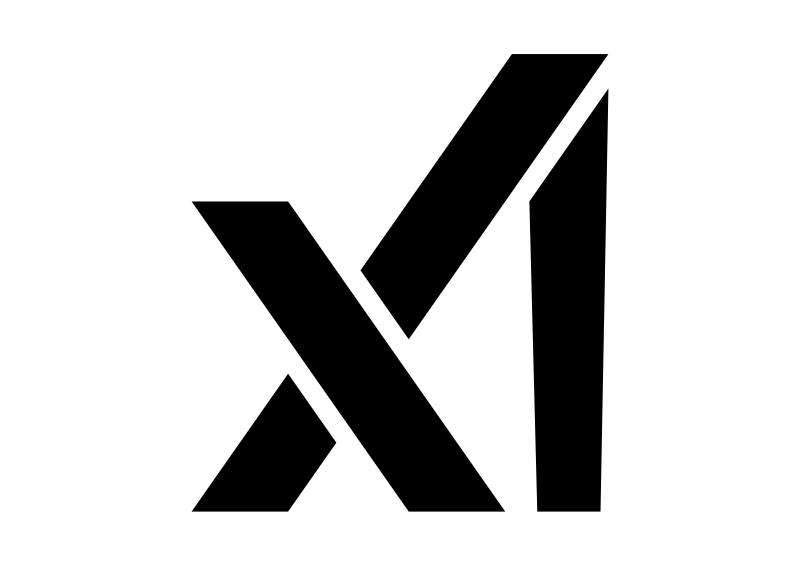}}\hspace{0.2em}xAI}
\newcommand{\anthropicsmall}{\raisebox{-0.25\height}{\includegraphics[height=1.2em]{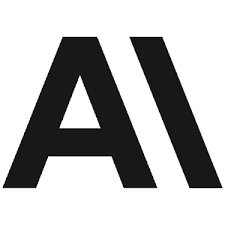}}\hspace{0.2em}Anthropic}
\newcommand{\googlesmall}{\raisebox{-0.25\height}{\includegraphics[height=1.2em]{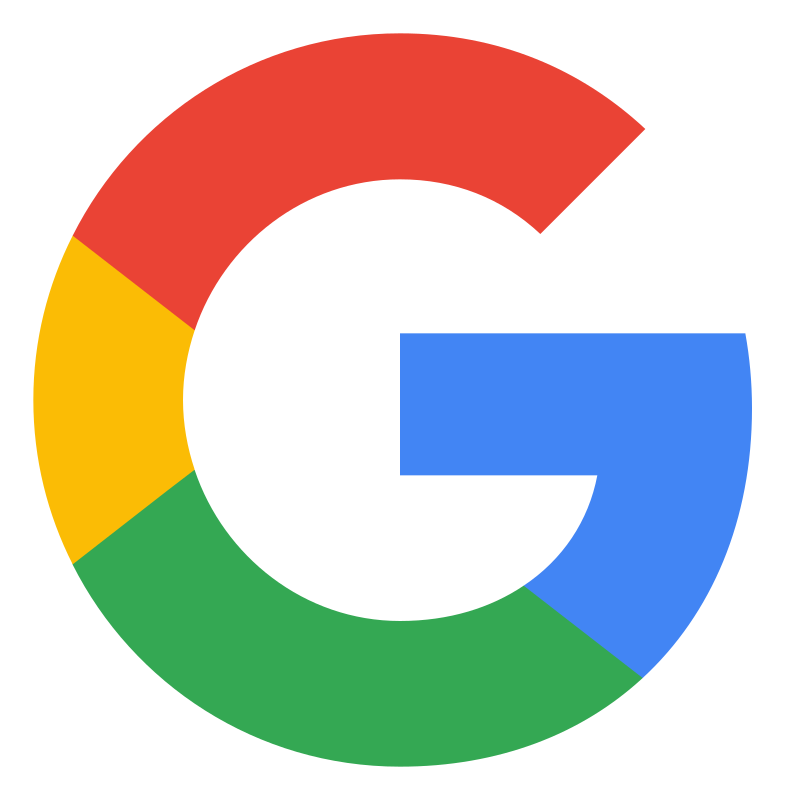}}\hspace{0.2em}Google}

\newcolumntype{P}[1]{>{\raggedright\arraybackslash}p{#1}}

\newmdenv[
  linecolor=black,
  linewidth=.6pt,
  leftmargin=0pt,
  rightmargin=0pt,
  innerleftmargin=6pt,
  innerrightmargin=6pt,
  innertopmargin=6pt,
  innerbottommargin=6pt,
  roundcorner=0pt,
  skipabove=6pt,
  skipbelow=6pt,
  middlelinewidth=0pt
]{qbox}

\makeatletter
\newcommand{\@trackname}{Main Track}
\makeatother

\title{Reasoning Models Ace the CFA Exams}

\author{
  Jaisal Patel$^1$, Yunzhe Chen$^2$\thanks{Co-primary author.} , Kaiwen He$^3$, Keyi Wang$^3$, 
  David Li$^4$, \\
  \textbf{Kairong Xiao$^5$, Xiao-Yang Liu Yanglet$^3$}\thanks{Corresponding author.} \\
  $^1$Rensselaer Polytechnic Institute, Troy, NY 12180 \\
  $^2$University of North Carolina at Chapel Hill, Chapel Hill, NC 27599 \\
  $^3$SecureFinAI Lab, Columbia University, New York, NY 10027 \\
  $^4$Department of Mathematics, Columbia University, New York, NY 10027 \\
  $^5$Business School, Columbia University, New York, NY 10027 \\
  Emails: XL2427@columbia.edu \\
}

\begin{document}
\raggedbottom

\maketitle

\begin{abstract}

Previous research has reported that large language models (LLMs) demonstrate poor performance on the Chartered Financial Analyst (CFA) exams. However, recent reasoning models have achieved strong results on graduate-level academic and professional examinations across various disciplines. In this paper, we evaluate state-of-the-art reasoning models on a set of mock CFA exams consisting of 980 questions across three Level I exams, two Level II exams, and three Level III exams. Using the same pass/fail criteria from prior studies, we find that most models clear all three levels. The models that pass, ordered by overall performance, are Gemini 3.0 Pro, Gemini 2.5 Pro, GPT-5, Grok 4, Claude Opus 4.1, and DeepSeek-V3.1. Specifically, Gemini 3.0 Pro achieves a record score of 97.6\% on Level I. Performance is also strong on Level II, led by GPT-5 at 94.3\%. On Level III, Gemini 2.5 Pro attains the highest score with 86.4\% on multiple-choice questions while Gemini 3.0 Pro achieves 92.0\% on constructed-response questions.

\end{abstract}

\section{Introduction}

  The evaluation of large language models (LLMs) on high-stakes, domain-specific examinations has become a critical measure of their advancing capabilities. While impressive results on major benchmarks like the medical USMLE, the mathematical AIME, and the legal Uniform Bar Exam have demonstrated LLMs' broad knowledge and reasoning, these assessments often test knowledge retrieval and logical deduction in isolation. In contrast, the finance domain \cite{liu2023fingpt,xie2024finben,yanglet2025multimodal} requires the simultaneous application of precise numerical calculations, qualitative analysis, and ethical judgment.
  
  The Chartered Financial Analyst (CFA) certification is a globally recognized qualification for investment and financial professionals. The CFA program is structured into three levels that test an evolving hierarchy of skills and formats: Level I tests foundational knowledge through individual multiple-choice questions (MCQs); Level II tests application and analysis via case-based multiple-choice item sets (vignettes); and Level III tests complex synthesis and portfolio construction across specialized pathways using a combination of item sets and constructed-response questions (CRQs). This structure provides a detailed method for assessing LLM capabilities, allowing for an evaluation of foundational knowledge, application, and complex synthesis.
  
  Beginning in 2023, research on CFA exams with LLMs has progressed from demonstrating poor performance to achieving passing scores. \citep{callanan2023can} reported that ChatGPT (GPT-3.5-turbo) failed Levels I and II, while GPT-4 passed Level I and failed Level II. In 2024, \citep{mahfouz-etal-2024-state} found that Claude 3 Opus and GPT-4o pass Levels I and II. More recently, \citep{shetty2025advanced} showed that o4-mini, Gemini 2.5 Pro, and DeepSeek-R1 passed Level III. While these studies demonstrate rapid improvement, a single comprehensive evaluation of the recent generation of reasoning models across all three CFA levels remains absent.

  In this paper, we first reproduce the results of general LLMs from \cite{callanan2023can, mahfouz-etal-2024-state} on a set of mock exams for all levels of the CFA (three Level I, two Level II, and three Level III), consisting of a total of 980 questions, using the same settings and pass/fail criteria to establish a baseline. Second, we evaluate the current state-of-the-art reasoning models, including GPT-5, Gemini 3.0 Pro, DeepSeek-V3.1, and Grok 4, alongside predecessors, such as Gemini 2.5 Pro and Claude Opus 4.1, on the same set of mock exams, settings, and criteria. We find that most models meet the passing threshold across all three levels, as shown in \Cref{tab:pass_fail_setB_sameformat}, with detailed accuracy metrics provided in \Cref{tab:overall_perf_setB_sameformat}.
  
\begin{table}[H]
\centering
\renewcommand{\arraystretch}{0.95}
\caption{Pass/Fail outcomes for LLMs on mock CFA exams. Models are ranked by their average accuracy across all three levels. \black{Black} for \cite{callanan2023can,mahfouz-etal-2024-state} and \blue{blue} for ours.}
\label{tab:pass_fail_setB_sameformat}
\begin{adjustbox}{max width=\textwidth}
\begin{tabular}{c |cc | ccc}
\toprule
\bfseries Model Producer & \bfseries Ranking & \bfseries Model & \bfseries Level I & \bfseries Level II & \bfseries Level III \\
\midrule
\multirow{5.5}{*}{\openaismall}
  & 9 &  \multirow{1}{*}{ChatGPT \cite{ouyang2022training}}
      & \black{Fail}/\blue{Fail} & \black{Fail}/\blue{Fail} & \blue{Fail} \\
\cmidrule(lr){2-6}
  & 8 & \multirow{1}{*}{GPT-4 \cite{achiam2023gpt}}
      & \black{Pass}/\blue{Pass} & \black{Pass}/\blue{Pass} & \blue{Fail} \\
  \cmidrule(lr){2-6}  
  & 7 & \multirow{1}{*}{GPT-4o \cite{hurst2024gpt}}
      & \black{Pass}/\blue{Pass} & \black{Pass}/\blue{Pass} & \blue{Pass} \\
\cmidrule(lr){2-6}
  & 3 & \multirow{1}{*}{GPT-5 \cite{openai2025gpt5}}
      &  \blue{Pass} &  \blue{Pass} & \blue{Pass} \\
\midrule
\multirow{2.5}{*}{\googlesmall}
  & 2 & \multirow{1}{*}{Gemini 2.5 Pro \cite{comanici2025gemini}}
      &  \blue{Pass} &  \blue{Pass} &  \blue{Pass} \\
\cmidrule(lr){2-6}
  & 1 & \multirow{1}{*}{Gemini 3.0 Pro}
      &  \blue{Pass} &  \blue{Pass} &  \blue{Pass} \\

\midrule
\multirow{1}{*}{\deepseeksmall}
  & 6 & \multirow{1}{*}{DeepSeek-V3.1 \cite{deepseekai2024deepseekv3technicalreport}}
      & \blue{Pass} & \blue{Pass} & \blue{Pass} \\
\midrule
\multirow{1}{*}{\xaismall}
  & 4 & \multirow{1}{*}{Grok 4 \cite{xai2025grok4}}
      &  \blue{Pass} &  \blue{Pass} & \blue{Pass} \\
\midrule
\multirow{1}{*}{\anthropicsmall}
 & 5 & \multirow{1}{*}{Claude Opus 4.1 \cite{anthropic2025claude}}
      &  \blue{Pass} &  \blue{Pass} & \blue{Pass} \\
\bottomrule
\end{tabular}
\end{adjustbox}
\vspace{-0.15in}
\end{table}

\section{Mock CFA Exam Dataset}
\label{headings}

\subsection{Question Set}
We compile a set of mock CFA exams across all three levels, with a total of 980 questions. The Level I set consists of three exams totaling 540 independent MCQs (180 per exam). The Level II set consists of two exams totaling 176 MCQs (88 per exam), organized into 22 item sets per exam (4 questions per set). The Level III set consists of three exams totaling 264 questions (88 per exam); each exam follows a hybrid format of 11 item sets (totaling 44 MCQs) and 11 constructed-response case studies (totaling 44 CRQs). Although the precise number and point-weighting of constructed-response questions vary in official CFA exams, these mock exams adhere to a standard, representative structure.

\enlargethispage{2cm}
\begin{figure}[H]
\centering
\includegraphics[width=0.96\textwidth, trim=0cm 0cm 0cm 0.055cm, clip]{images/Example_Figure.pdf}
\caption{Sample mock CFA exam questions by level. Cases are shown in \textcolor[HTML]{0632b5}{blue}, questions in \textcolor[HTML]{c4165b}{red}, and answer choices in \textcolor[HTML]{096f52}{green}. Examples are illustrative and not actual exam content.}
\label{example_cfa_questions}
\vspace{-0.2in}
\end{figure}

\clearpage
\subsection{Composition and Reproduction Validity}
\textbf{Data sources}. We compile the mock exam dataset from two primary sources: \blacklink{https://www.cfainstitute.org/programs/cfa-program/candidate-resources/practice-pack}{the official CFA Institute Practice Pack} and \blacklink{https://app.analystprep.com}{AnalystPrep}. For Levels I and II, we use the CFA Practice Pack from 2024 and 2025, respectively. For Level III, we use AnalystPrep Mock Exams from 2025. 

In contrast, prior studies relied on different sources: \cite{callanan2023can} and \cite{mahfouz-etal-2024-state} used AnalystPrep Mock Exams from 2023 and 2024, respectively. Exact data replication is precluded because the questions in \cite{callanan2023can} remain undisclosed, and the datasets in \cite{mahfouz-etal-2024-state} reflect a superseded curriculum. Specifically, the 2024 curriculum update emphasized conceptual application for Levels I and II by shifting foundational calculations to prerequisite readings and substantially revising topics such as Corporate Issuers and Fixed Income, while the 2025 update introduced specialized Pathways for Level III. Furthermore, relying on older datasets increases the risk of benchmark contamination, where model performance reflects training data contamination rather than reasoning capability. Therefore, we ensure validity by utilizing materials that match the current examination standard, preserving the difficulty and relevance of the evaluation.

\textbf{Topic distribution}. The Levels I and II exams in the mock exam dataset cover all ten standard topics. For Level III, the structure aligns with the 2025 curriculum, covering six key areas: Asset Allocation (15–20\%), Portfolio Construction (15–20\%), Performance Measurement (5–10\%), Derivatives \& Risk Management (10–15\%), Ethical Standards (10–15\%), and the specialized Pathways (30–35\%) in either Portfolio Management, Private Markets, or Private Wealth.

To validate representativeness, we compare our topic weight distribution with \cite{callanan2023can} and \cite{mahfouz-etal-2024-state} in \Cref{tab:topic_weights_comparison}. Note that to ensure consistent comparison across studies, we map the topics to the high-level functional domains established in \cite{mahfouz-etal-2024-state}, distributing them across Ethical Standards, Investment Tools, Asset Classes, and Portfolio Management. The exact breakdown of topics and question counts covered in the Level III mock exams is provided in \cref{tab:l3_errors_by_topic_styled} in the appendix.

\begin{table}[H]
\centering
\small
\setlength{\tabcolsep}{3pt} 

\begin{minipage}{\textwidth}
\centering

\caption{Comparison of mock exam topic weights (percentage) across studies.}
\label{tab:topic_weights_comparison}

\begin{tabular}{l ccc ccc ccc}
\toprule
 & \multicolumn{3}{c}{\textbf{Level I}} 
 & \multicolumn{3}{c}{\textbf{Level II}} 
 & \multicolumn{3}{c}{\textbf{Level III}} \\
\cmidrule(lr){2-4} \cmidrule(lr){5-7} \cmidrule(lr){8-10}

\textbf{Topic Area} 
 & \textbf{\cite{callanan2023can}} & \textbf{\cite{mahfouz-etal-2024-state}} & \textbf{\blue{This Work}} 
 & \textbf{\cite{callanan2023can}} & \textbf{\cite{mahfouz-etal-2024-state}} & \textbf{\blue{This Work}} 
 & \textbf{\cite{callanan2023can}} & \textbf{\cite{mahfouz-etal-2024-state}} & \textbf{\blue{This Work}} \\
\midrule
\textbf{Ethical Standards} & 16.1\% & 16.0\% & \blue{15.0\%} & 11.4\% & 11.0\% & \blue{13.6\%} & - & 9.0\% & \blue{13.6\%} \\
\midrule
\textbf{Investment Tools} & 39.2\% & 39.0\% & \blue{35.4\%} & 43.1\% & 43.0\% & \blue{34.0\%} & - & 0.0\% & \blue{0.0\%} \\
\hspace{1em}Quantitative Methods & 9.8\% & 10.0\% & \blue{8.0\%} & 10.2\% & 10.0\% & \blue{6.8\%} & - & - & \blue{-} \\
\hspace{1em}Economics & 9.7\% & 10.0\% & \blue{7.6\%} & 6.8\% & 7.0\% & \blue{6.8\%} & - & - & \blue{-} \\
\hspace{1em}Financial Reporting & 13.7\% & 14.0\% & \blue{12.2\%} & 15.9\% & 16.0\% & \blue{13.6\%} & - & - & \blue{-} \\
\hspace{1em}Corporate Issuers & 6.0\% & 5.0\% & \blue{7.6\%} & 10.2\% & 10.0\% & \blue{6.8\%} & - & - & \blue{-} \\
\midrule
\textbf{Asset Classes} & 38.0\% & 38.0\% & \blue{39.1\%} & 36.3\% & 37.0\% & \blue{40.8\%} & - & 32.0\% & \blue{24.3\%} \\
\hspace{1em}Equity Investments & 15.9\% & 16.0\% & \blue{12.4\%} & 13.6\% & 14.0\% & \blue{13.6\%} & - & - & \blue{-} \\
\hspace{1em}Fixed Income & 10.3\% & 10.0\% & \blue{12.0\%} & 12.5\% & 13.0\% & \blue{13.6\%} & - & - & \blue{-} \\
\hspace{1em}Derivatives & 3.2\% & 3.0\% & \blue{8.0\%} & 6.8\% & 7.0\% & \blue{6.8\%} & - & - & \blue{-} \\
\hspace{1em}Alternative Investments & 8.6\% & 9.0\% & \blue{6.7\%} & 3.4\% & 3.0\% & \blue{6.8\%} & - & - & \blue{-} \\
\midrule
\textbf{Portfolio Management} & 6.7\% & 7.0\% & \blue{10.6\%} & 9.1\% & 9.0\% & \blue{11.4\%} & - & 59.0\% & \blue{-} \\
\midrule
\textbf{Pathways} & - & - & \blue{-} & - & - & \blue{-} & - & - & \blue{62.1\%}\textsuperscript{$\dagger$} \\

\midrule
\midrule
\textbf{\#Mock exams} & 5 & 2 & \blue{3} & 2 & 2 & \blue{2} & - & 2 & \blue{3} \\
\midrule
\textbf{\#Questions} & 180 & 180 & \blue{180} & 88 & 88 & \blue{88} & - & 44* & \blue{88} \\
\bottomrule
\end{tabular}

\vspace{0.5em}
\raggedright
\footnotesize
\textsuperscript{*} \cite{mahfouz-etal-2024-state} reports 44 questions for Level III, likely referring to a single session. Our evaluation uses full-length exams. \\
\textsuperscript{$\dagger$} Level III mock exams in this work are the only ones using the 2025 updated curriculum, which introduces Pathways. This allocation is comparable to the 59.0\% Portfolio Management weight in Level III reported by \cite{mahfouz-etal-2024-state}.
\end{minipage}
\end{table}

As shown, the topic distribution broadly mirrors prior datasets. For Level I, the maximum deviation in topic weights is within 4.0 percentage points of both \cite{callanan2023can} and \cite{mahfouz-etal-2024-state}. For Level II, we observe a notable shift in Investment Tools (34.0\% vs. 43.0\% and 43.1\% in prior work), consistent with the reduced emphasis on foundational tools. For Level III, comparison with \cite{mahfouz-etal-2024-state} shows a reallocation of weight from Asset Classes to Ethical Standards and Pathways in the mock exams we use. Despite these adjustments, Investment Tools and Asset Classes constitute the majority of the curriculum for Levels I and II, whereas Level III retains its concentration on portfolio application. The functional focus of each level remains consistent across datasets, ensuring a comparable scope for reproduction.

\textbf{Structural characteristics}. We further validate our dataset by comparing the characteristics of Level I and Level II questions with those reported in \cite{callanan2023can}. As shown in \Cref{tab:question_stats_comparison}, we observe a notable shift in question composition; specifically, our Level I dataset exhibits a lower density of calculation-based questions in Quantitative Methods (39.5\% vs. 70.5\%) and Economics (12.2\% vs. 50.6\%). This difference, once again, reflects the updated curriculum, which places greater emphasis on conceptual application, as well as variability in topic weighting across different mock exams. By validating models against the active curriculum, we ensure that the evaluation targets relevant professional requirements while preserving the difficulty required for a valid reproduction. This complexity is supported by an increase in information density, where our dataset features significantly longer average prompt lengths across both levels compared to prior work.

\begin{table}[H]\small
\centering
\caption{Levels I and II question characteristics by topics: percentage of questions with numerical calculation, average number of tables per question, and average prompt length in tokens. 
\black{Black} for \cite{callanan2023can} and  
\blue{blue} for ours.}
\label{tab:question_stats_comparison}

\setlength{\tabcolsep}{6pt}
\renewcommand{\arraystretch}{0.95}
\begin{adjustbox}{max width=\textwidth}
\begin{tabular}{l|ccc|ccc}
\toprule
        & \multicolumn{3}{c|}{\bfseries Level I} 
        & \multicolumn{3}{c}{\bfseries Level II} \\ 
\cmidrule(lr){2-4}\cmidrule(lr){5-7}
\bfseries Topic 
        & Calculation & \#Tab. & Length 
        & Calculation & \#Tab. & Length \\
\midrule
Ethics                 & \black{0.7\%}/\blue{0.0\%}    & \black{0.01}/\blue{0.06} & \black{125}/\blue{158}    & \black{0.0\%}/\blue{0.0\%} & \black{0.00}/\blue{0.00} & \black{1013}/\blue{1162} \\
Quantitative Methods          & \black{70.5\%}/\blue{39.5\%} & \black{0.26}/\blue{0.26} & \black{131}/\blue{148}    & \black{27.8\%}/\blue{50.0\%} & \black{0.00}/\blue{1.67} & \black{1256}/\blue{1197} \\
Economics                      & \black{50.6\%}/\blue{12.2\%} & \black{0.25}/\blue{0.12} & \black{121}/\blue{147}    & \black{66.7\%}/\blue{41.7\%} & \black{2.00}/\blue{1.33} & \black{1115}/\blue{1020} \\
Financial Reporting            & \black{57.7\%}/\blue{33.3\%} & \black{0.35}/\blue{0.32} & \black{151}/\blue{152}    & \black{53.6\%}/\blue{33.3\%} & \black{2.79}/\blue{1.50} & \black{1383}/\blue{1072} \\
Corporate Issuers              & \black{59.3\%}/\blue{19.5\%} & \black{0.28}/\blue{0.20} & \black{120}/\blue{146}    & \black{44.4\%}/\blue{58.3\%} & \black{1.67}/\blue{2.00} & \black{930}/\blue{1135}  \\
Equity Investments             & \black{52.5\%}/\blue{28.4\%} & \black{0.19}/\blue{0.27} & \black{112}/\blue{150}    & \black{45.8\%}/\blue{58.3\%} & \black{1.00}/\blue{1.50} & \black{1053}/\blue{1048} \\
Fixed Income                   & \black{43.0\%}/\blue{30.8\%} & \black{0.06}/\blue{0.15} & \black{87}/\blue{151}     & \black{50.0\%}/\blue{58.3\%} & \black{1.45}/\blue{1.17} & \black{779}/\blue{1089}  \\
Derivatives                    & \black{20.7\%}/\blue{20.9\%} & \black{0.00}/\blue{0.07} & \black{65}/\blue{159}     & \black{75.0\%}/\blue{58.3\%} & \black{2.00}/\blue{1.00} & \black{816}/\blue{1073}  \\
Alternative Investments        & \black{36.4\%}/\blue{13.9\%} & \black{0.06}/\blue{0.11} & \black{85}/\blue{157}     & \black{66.7\%}/\blue{50.0\%} & \black{2.00}/\blue{2.33} & \black{840}/\blue{1212}  \\
Portfolio Management           & \black{38.3\%}/\blue{24.6\%} & \black{0.18}/\blue{0.19} & \black{110}/\blue{152}    & \black{56.3\%}/\blue{25.0\%} & \black{2.13}/\blue{1.40} & \black{1077}/\blue{1100} \\
\midrule
\textbf{Overall}               & \black{42.4\%}/\blue{22.0\%} & \black{0.17}/\blue{0.18} & \black{116}/\blue{152}    & \black{45.5\%}/\blue{40.9\%} & \black{1.47}/\blue{1.30} & \black{1058}/\blue{1111} \\
\bottomrule
\end{tabular}
\end{adjustbox}
\vspace{-0.2in}
\end{table}

\section{Evaluation Methodology}
\subsection{Experimental Setup}

\textbf{LLMs}. We evaluate three groups of models: i) baselines used in \cite{callanan2023can} (ChatGPT, GPT-4); ii) the subsequent model from \cite{mahfouz-etal-2024-state} (GPT-4o); and iii) the state-of-the-art reasoning models (GPT-5, Gemini 3.0 Pro, DeepSeek-V3.1, Grok 4) alongside predecessors (Gemini 2.5 Pro, Claude Opus 4.1). The exact model identifiers and snapshots (date-stamped versions of models) are provided in \Cref{tab:model_snapshots}. 

\begin{table}[H]\small
\centering
\caption{Specific model identifiers and version snapshots used in our evaluation.}
\label{tab:model_snapshots}
\setlength{\tabcolsep}{6pt}
\renewcommand{\arraystretch}{1.05}
\begin{tabular}{lllc}
\toprule
\textbf{Provider} & \textbf{Model} & \textbf{Identifier} & \textbf{Snapshot Date} \\
\midrule
\multicolumn{4}{l}{\textit{Baselines reproduced from \cite{callanan2023can} and \cite{mahfouz-etal-2024-state}}} \\
OpenAI & ChatGPT (GPT-3.5 Turbo) & \texttt{gpt-3.5-turbo} & 25 Jan 2024 \\
OpenAI & GPT-4 & \texttt{gpt-4} & 13 Jun 2023 \\
OpenAI & GPT-4o & \texttt{gpt-4o-2024-08-06} & 06 Aug 2024 \\
\midrule
\multicolumn{4}{l}{\textit{Reasoning Models (This Work)}} \\
OpenAI & GPT-5 & \texttt{gpt-5-preview} & 07 Aug 2025 \\
Google & Gemini 2.5 Pro & \texttt{gemini-2.5-pro} & 17 Jun 2025 \\
Google & Gemini 3.0 Pro & \texttt{gemini-3-pro-preview} & 18 Nov 2025 \\
xAI & Grok 4 & \texttt{grok-4} & 09 Jul 2025 \\
Anthropic & Claude Opus 4.1 & \texttt{claude-4.1-opus} & 08 Aug 2025 \\
DeepSeek & DeepSeek-V3.1 & \texttt{deepseek-v3.1} & 28 May 2025 \\
\bottomrule
\end{tabular}
\end{table}

\textbf{Model parameters}. To ensure reproducibility and comparability across models, we use provider-default parameters for all API calls, with the temperature set to 0 to minimize generation randomness. Exceptions were made for models where the temperature is not configurable (e.g., GPT-5). Note that a zero-temperature setting may impact the optimal reasoning capability of models. Due to variability across runs, we report results as the average score $\pm$ standard deviation.

\textbf{Prompting}. We evaluate model performance under two distinct prompting conditions. The exact prompt templates for each setting are provided in \cref{app:prompts}. 
\begin{itemize}[leftmargin=*, label=$\bullet$]
    \item \textbf{Zero-Shot (ZS):} The model is presented with the question context and instructed to output the final answer directly. 
    \item \textbf{Chain-of-Thought (CoT):} We use a Zero-Shot Chain-of-Thought approach, instructing the model to "think step-by-step" and "explain your reasoning" before generating the final answer.
\end{itemize}

\subsection{Evaluation Metrics}

\textbf{Scoring}. Each MCQ consists of three options, with exactly one correct answer. We report accuracy as the number of correct responses divided by the total question count. For CRQs, we employ o4-mini as an automated evaluator. We provide the model with the case context, question, reference answer, candidate response, and the AnalystPrep grading rubric. The specific prompt structure is detailed in \cref{app:prompts_automated_grading}. Final CRQ scores are reported as a percentage, calculated as the total points awarded divided by the maximum possible score (132 points per exam).

\textbf{Pass/Fail criteria}. We adopt the following passing thresholds from prior studies to ensure consistent evaluation:

\begin{enumerate}[leftmargin=*, label=\textbf{Level \Roman*.}]    
    \item Pass if the score is $\ge 60\%$ in every individual topic \textit{and} $\ge 70\%$ overall \cite{callanan2023can}.

    \item Pass if the score is $\ge 50\%$ in every individual topic \textit{and} $\ge 60\%$ overall \cite{callanan2023can}.

    \item Pass if the average of the MCQ and CRQ scores is $\ge 63\%$ \cite{300hourscfa}.

\end{enumerate}  

\vspace{-0.1in}
\section{Experiment Results}

\label{others}

\subsection{Reproduction of Previous Work}
We verify existing results from previous works by re-evaluating ChatGPT, GPT-4, and GPT-4o models under identical settings. Comparing our obtained results with those from \cite{callanan2023can, mahfouz-etal-2024-state}, such a reproduction serves as a validated baseline for evaluating reasoning models.

Our results align with \cite{callanan2023can}, confirming that general-purpose LLMs perform poorly across all levels. ChatGPT consistently fails to meet passing criteria, achieving scores of 58.9\%-68.4\% on CFA Level I and 43.8\%-48.3\% on Level II. GPT-4 demonstrates stronger performance, scoring 73.3\%–80.9\% on Level I and 55.7\%–69.9\% on Level II, yet still fails to clear the Level II threshold in Zero-Shot settings. We find CoT prompting leads to substantial performance gains for GPT-4 (7.6–14.2 percentage points) and moderate gains for ChatGPT (4.5–5.5 percentage points) without changing the pass/fail outcome.

Consistent with \cite{mahfouz-etal-2024-state}, we find that GPT-4o reliably passes Levels I and II. Our evaluation shows scores of 90.6\% on Level I and 73.9\% on Level II under the CoT setting, closely tracking the 88.1\% and 76.7\% reported in their evaluation. A notable divergence occurs at Level III, whereas \cite{mahfouz-etal-2024-state} reported a failing constructed-response score of 46.2\%, our evaluation shows a passing score of 66.7\%. This difference likely stems from the evolution of the mock exam datasets (2024 vs. 2025) and our use of the latest stable model snapshot (\texttt{gpt-4o-2024-08-06}), which offers improvements over the unspecified release evaluated in \cite{mahfouz-etal-2024-state}.

\subsection{Evaluation of State-of-the-Art Reasoning Models}
We evaluate the current state-of-the-art reasoning models, specifically GPT-5, Gemini 3.0 Pro, DeepSeek-V3.1, and Grok 4, alongside a subset of predecessor models including Gemini 2.5 Pro, and Claude Opus 4.1.

Previous papers \cite{callanan2023can, mahfouz-etal-2024-state} reported that LLMs at the time were unable to pass all three levels of the CFA exam. Our results demonstrate that this is no longer the case. As shown in \Cref{tab:overall_perf_setB_sameformat}, this entire set of reasoning models not only passes all three CFA levels but also achieves nearly perfect scores in Levels I and II. Gemini 3.0 Pro achieves the highest score on Level I with{97.6\%} (ZS), while GPT-5 leads on Level II with {94.3\%} (ZS). 

On the Level II exam, the predecessor model {Gemini 2.5 Pro} scores the highest accuracy on multiple-choice questions ({86.4\%}), and the newer {Gemini 3.0 Pro} demonstrates a significant advantage on constructed-response questions, scoring {92.0\%} compared to 82.8\% for Gemini 2.5 Pro. These results indicate that reasoning models surpass the expertise required of entry-level to mid-level financial analysts and may achieve senior-level financial analyst proficiency in the future. These results indicate that while this class of models has mastered the codified knowledge of Levels I and II, the latest state-of-the-art iterations are specifically extending capabilities in the complex synthesis required for Level III.

\begin{table}[H]\small
\centering
\renewcommand{\arraystretch}{0.6}
\caption{Overall performance of models on mock CFA exams (Accuracy) in zero-shot (ZS) and chain-of-thought (CoT). \black{Black} for \cite{callanan2023can,mahfouz-etal-2024-state} and \blue{blue} for ours.}
\label{tab:overall_perf_setB_sameformat}
\begin{adjustbox}{max width=\textwidth}
\begin{tabular}{cc| cccc}
\toprule
\bfseries Model & \bfseries Setting & \bfseries Level I & \bfseries Level II & \bfseries Level III MCQ & \bfseries Level III CRQ \\
\midrule
\multirow{2}{*}{ChatGPT \cite{ouyang2022training}}
    & ZS  & \black{58.8$\pm$0.2}/\blue{58.9$\pm$1.0} & \black{46.6$\pm$0.6}/\blue{43.8$\pm$3.5} & \blue{54.5$\pm$1.9} & \blue{44.4$\pm$1.9} \\
  & CoT & \black{58.0$\pm$0.2}/\blue{64.4$\pm$2.5} & \black{47.2$\pm$0.3}/\blue{48.3$\pm$1.0} & \black{44.2$\pm$6.0}/\blue{52.3$\pm$3.5} & \black{17.4$\pm$2.1}/\blue{44.4$\pm$1.5} \\
  \cmidrule(lr){1-6}
\multirow{2}{*}{GPT-4 \cite{achiam2023gpt}} 
    & ZS  & \black{73.2$\pm$0.2}/\blue{73.3$\pm$1.0} & \black{57.4$\pm$1.5}/\blue{55.7$\pm$2.5} & \blue{59.8$\pm$1.9} & \blue{57.8$\pm$1.7} \\
  & CoT & \black{74.0$\pm$0.2}/\blue{80.9$\pm$3.3} & \black{61.4$\pm$0.9}/\blue{69.9$\pm$4.3} & \blue{65.9$\pm$3.8} & \blue{58.1$\pm$3.2} \\
  \cmidrule(lr){1-6}  
\multirow{2}{*}{GPT-4o \cite{hurst2024gpt}}
    & ZS  & \blue{80.0$\pm$0.0} & \blue{71.6$\pm$1.4} & \blue{63.6$\pm$1.7} & \blue{62.8$\pm$5.6} \\
  & CoT & \black{88.1$\pm$0.3}/\blue{90.6$\pm$1.0} & \black{76.7$\pm$0.7}/\blue{73.9$\pm$5.2} & 
  \black{63.4$\pm$4.2}/\blue{68.9$\pm$1.0} & \black{46.2$\pm$3.3}/\blue{66.7$\pm$1.8} \\
\cmidrule(lr){1-6}
\multirow{2}{*}{GPT-5 \cite{openai2025gpt5}}
    & ZS  &  \blue{96.1$\pm$1.0} &  \blue{\textbf{94.3$\pm$2.9}} & \blue{73.5$\pm$ 2.5} & \blue{71.8$\pm$0.4} \\
  & CoT &  \blue{{96.7$\pm$1.9}} &  \blue{92.6$\pm$1.4} & \blue{75.0$\pm$3.5} & \blue{71.8$\pm$1.9} \\
\midrule
\midrule
\multirow{2}{*}{Gemini 2.5 Pro \cite{comanici2025gemini}}
    & ZS  &  \blue{95.7$\pm$0.0} &  \blue{92.6$\pm$1.4} & \blue{84.1$\pm$1.9} & \blue{78.2$\pm$3.0} \\
  & CoT &  \blue{96.1$\pm$1.7} &  \blue{92.6$\pm$1.4} & \blue{\textbf{86.4$\pm$3.8}} & \blue{{82.8$\pm$1.5}} \\
\cmidrule(lr){1-6}
\multirow{2}{*}{Gemini 3.0 Pro}
    & ZS  &  \textbf{\blue{97.6$\pm$0.0}} &  \blue{{93.2$\pm$0.0}} & \blue{81.8$\pm$0.0} & \blue{86.6$\pm$1.2} \\
  & CoT &  \blue{{97.0$\pm$0.5}} &  \blue{92.0$\pm$0.0} & \blue{80.3$\pm$0.0} & \textbf{\blue{92.0$\pm$3.5}} \\
\midrule
\midrule
\multirow{2}{*}{DeepSeek-V3.1 \cite{deepseekai2024deepseekv3technicalreport}}
    & ZS  & \blue{90.9$\pm$1.0} & \blue{85.2$\pm$2.5} & \blue{81.1$\pm$1.7} & \blue{70.8$\pm$4.3} \\
  & CoT & \blue{91.3$\pm$1.0} & \blue{85.8$\pm$1.4} & \blue{81.8 $\pm$0.0} & \blue{72.0$\pm$2.6} \\
\midrule
\midrule
\multirow{2}{*}{Grok 4 \cite{xai2025grok4}}
    & ZS  &  \blue{94.8$\pm$1.0} &  \blue{85.2$\pm$2.5} & \blue{78.0$\pm$1.0} & \blue{71.2$\pm$2.6} \\
  & CoT &  \blue{95.9$\pm$0.0} &  \blue{86.4 $\pm$1.4} & \blue{78.0$\pm$1.7} & \blue{80.2$\pm$3.3} \\
\midrule
\midrule
\multirow{2}{*}{Claude Opus 4.1 \cite{anthropic2025claude}}
    & ZS  &  \blue{94.6$\pm$1.0} &  \blue{89.8$\pm$1.4} & \blue{75.0$\pm$1.0} & \blue{73.4$\pm$3.1} \\
  & CoT &  \blue{94.8$\pm$1.0} &  \blue{89.8$\pm$4.3} & \blue{74.2$\pm$3.3} & \blue{79.0$\pm$4.8} \\
\bottomrule
\end{tabular}
\end{adjustbox}
\vspace{-0.2in}
\end{table}

\subsection{Performance Analysis}
\textbf{Impact of prompting strategy}. We observe a distinct divergence in the efficacy of Chain-of-Thought (CoT) prompting across model generations. For baseline models, CoT provides substantial gains on Levels I and II, improving GPT-4 accuracy by 7.6–14.2 percentage points and ChatGPT by 4.5–5.5 percentage points. This suggests that explicit reasoning steps are critical for earlier architectures to bridge the gap between knowledge recall and application.

In contrast, reasoning models exhibit inconsistent responses to explicit prompting on multiple-choice questions across all levels. While Grok 4 shows standard improvements, Gemini 3.0 Pro exhibits slight regressions under CoT settings for Level I (-0.6\%), Level II (-1.2\%), and Level III MCQs (-1.5\%). Similarly, GPT-5 shows performance drops on Level II (-1.7\%). However, this trend reverses for constructed-response questions, where CoT remains highly effective. For example, Gemini 3.0 Pro's performance on CRQs jumps from 86.6\% (ZS) to 92.0\% (CoT) and Claude Opus 4.1 from 73.4\% to 79.0\%. This suggests that while modern architectures may be approaching a performance ceiling for closed-ended tasks, explicit reasoning appears constructive for the synthesis required in open-ended tasks. We note, however, that this improvement on CRQs may partially reflect the verbosity bias of automated evaluators, a limitation further analyzed in \Cref{subsec:limitations_automation}.

\textbf{Topic-level performance shifts}. Our results also show a shift in difficulty distribution compared to \cite{mahfouz-etal-2024-state}, which identified quantitative domains as a primary weakness for LLMs. In our evaluation, advanced reasoning models appear to have overcome this bottleneck. For instance, GPT-5 and Grok 4 achieve near-zero error rates on {Quantitative Methods}, {Equity Investments}, and {Economics} across Levels I and II. Conversely, {Ethical and Professional Standards} remains a persistent challenge, exhibiting the highest relative error rates among the top-performing reasoning models (e.g., $\approx$17--21\% on Level II). The exact breakdown of errors by topic is provided in \cref{app:error}.

\textbf{Generational trade-offs}. We observe a distinct trade-off in the transition from Gemini 2.5 Pro to Gemini 3.0 Pro. While the newer model demonstrates superior capability in constructed-response questions (92.0\% vs. 82.8\%), it exhibits a slight regression on multiple-choice tasks (80.3\% vs. 86.4\% on Level III MCQs). 

\section{Limitations and Future Work}
\label{sec:limitations}
\subsection{CFA Exam Representation}
While our Level I and Level II evaluations use official CFA Institute material, our Level III dataset relies on third-party mock exams (AnalystPrep) to maintain consistency with \cite{mahfouz-etal-2024-state}. We acknowledge that third-party materials may differ from official examinations in vignette complexity, narrative depth, and distractor subtlety. Future work should prioritize official mock exams to maximize representativeness.

\subsection{Automated Scoring Validity}
\label{subsec:limitations_automation} 
For Level III constructed-response questions, we used o4-mini for automated grading based on standardized rubrics. This approach, while necessary for scalable evaluation, introduces potential measurement error. Research on LLM-based evaluation indicates a tendency toward a verbosity bias, where judges favor longer, comprehensive-sounding responses even if they lack specific technical precision. Furthermore, automated evaluators may fail to penalize subtle logical inconsistencies as strictly as human experts. The CRQ scores reported in this work should be interpreted as a model-based approximation. Future research requires validation by qualified CFA charterholders to establish a human-verified ground truth.

\subsection{Data Contamination Risks} 
A significant limitation of all LLM evaluations is the risk of training data contamination. While the mock exams used in this study are proprietary, paywalled, and relatively new, we cannot definitively rule out the possibility of indirect leakage. For example, paraphrased reconstructions or derivative discussions of these questions may appear in public training corpora. High performance could therefore partially reflect memorization rather than pure reasoning capability. Establishing a completely contamination-free evaluation environment remains an open challenge in the field.

\vspace{-0.1in}
\section{Conclusion}

In this paper, we present a comprehensive study of state-of-the-art reasoning models on mock CFA exams across all three levels, evaluating their performance against reproduced baselines from \cite{callanan2023can} and \cite{mahfouz-etal-2024-state} on a set of mock exams.

We find that top models achieve near-perfect scores on Level I (exceeding 97\%) and demonstrate high proficiency on Level II (over 94\%). Specifically, Gemini 3.0 Pro achieves a record score of 97.6\% on Level I, while GPT-5 leads on Level II with 94.3\%. On Level III, Gemini 2.5 Pro attains the highest score on multiple-choice questions with 86.4\%, while Gemini 3.0 Pro achieves 92.0\% on constructed-response questions.

These results suggest that current models have now largely mastered the codified knowledge base of Levels I and II. Furthermore, the substantial improvement in constructed-response performance by the latest reasoning generation indicates a growing capability for the complex synthesis required for Level III. Collectively, these findings establish new, unified performance baselines for future research.

\section*{Acknowledgement}
    
Keyi Wang and Xiao-Yang Liu Yanglet acknowledge the support from Columbia's SIRS and STAR Program, The Tang Family Fund for Research Innovations in FinTech, Engineering, and Business Operations.

\bibliographystyle{plain}
\bibliography{ref}

\newpage
\appendix

\section{Prompts and Instructions}
\label{app:prompts}

\subsection{Multiple Choice Questions (Level I, II, \& III MCQ)}

\begin{qbox}
\raggedright
\noindent\textbf{\textsc{System Prompts}}

\vspace{0.5em}
\textbf{Level I (Zero-Shot):}
``You are a CFA (chartered financial analyst) taking a test to evaluate your knowledge of finance. You will be given a question along with three possible answers (A, B, and C). Provide only the correct answer (A, B, or C) without any reasoning or explanation.''

\vspace{0.5em}
\textbf{Level II \& III MCQ (Zero-Shot with Context/Vignette):}
``You are a CFA (chartered financial analyst) taking a test to evaluate your knowledge of finance. You will be given a case description and a question along with three possible answers (A, B, and C). Provide only the correct answer (A, B, or C) without any reasoning or explanation.''

\vspace{0.5em}
\textbf{Level I (Chain-of-Thought):}
``You are a CFA (chartered financial analyst) taking a test to evaluate your knowledge of finance. You will be given a question along with three possible answers (A, B, and C). Before answering, think through the question step-by-step. Explain your reasoning, including any calculations. Indicate the correct answer (A, B, or C).''

\vspace{0.5em}
\textbf{Level II \& III MCQ (Chain-of-Thought with Context/Vignette):}
``You are a CFA (chartered financial analyst) taking a test to evaluate your knowledge of finance. You will be given a case description and a question along with three possible answers (A, B, and C). Before answering, think through the case step-by-step. Explain your reasoning, including any calculations. Indicate the correct answer (A, B, or C).''
\end{qbox}

\begin{qbox}
\raggedright
\noindent\textbf{\textsc{User Prompts}}

\vspace{0.5em}
\textbf{Level I Question:}
\begin{verbatim}
Question:
{question}
A. {a}
B. {b}
C. {c}
\end{verbatim}

\vspace{0.5em}
\textbf{Level II \& III Question:}
\begin{verbatim}
Case:
{context}

Question:
{question}
A. {a}
B. {b}
C. {c}
\end{verbatim}
\end{qbox}

\newpage

\subsection{Level III Constructed-Response Questions (CRQ)}

\begin{qbox}
\raggedright
\noindent\textbf{\textsc{System Prompts}}

\vspace{0.5em}
\textbf{Zero-Shot (ZS):}
``You are taking a test for the Chartered Financial Analyst (CFA) program designed to evaluate your knowledge of different topics in finance. You will be given a constructed-response question. Provide a clear, direct answer to the question.''

\vspace{0.5em}
\textbf{Chain-of-Thought (CoT):}
``You are taking a test for the Chartered Financial Analyst (CFA) program designed to evaluate your knowledge of different topics in finance. You will be given a constructed-response question. Think step-by-step and respond with your thinking and answer the question.''
\end{qbox}

\begin{qbox}
\raggedright
\noindent\textbf{\textsc{User Prompts}}

\begin{verbatim}
Case:
{context}

Question:
{question}

Answer:
\end{verbatim}
\end{qbox}

\newpage

\subsection{Automated Grading (Meta-Prompt)}
\label{app:prompts_automated_grading}

\begin{qbox}
\raggedright
\noindent\textbf{\textsc{Automated Grading Prompts}}

These prompts are used to grade CRQ responses.

\vspace{0.5em}
\textbf{System Prompt:}
``You are a CFA Level 3 examiner tasked with grading constructed-response answers. You will be provided with:
\begin{enumerate}[label=\arabic*., nolistsep]
    \item The complete vignette/case context
    \item The specific question
    \item The model answer with detailed explanation
    \item The grading rubric with specific criteria
    \item The student's answer (which may include reasoning process)
    \item The score range: minimum 0, maximum \{total\_points\} points
\end{enumerate}

Your task is to assign an integer score based on:
\begin{itemize}[nolistsep]
    \item Technical accuracy
    \item Reasoning quality
    \item Relevance to the question
    \item Communication clarity
    \item Strict adherence to the grading rubric
\end{itemize}

Requirements:
\begin{itemize}[nolistsep]
    \item Assign ONLY integer scores within the range 0--\{total\_points\}
    \item Follow the rubric criteria exactly for point allocation
    \item Base scoring on technical accuracy and rubric adherence
    \item Return format: "Score: X" followed by brief justification''
\end{itemize}

\vspace{0.5em}
\textbf{User Prompt:}
\begin{verbatim}
Please grade this student's answer strictly according to the rubric:

COMPLETE VIGNETTE/CASE:
{context}

QUESTION:
{question}

MODEL ANSWER:
{explanation}

GRADING RUBRIC:
{grading_rubric}

SCORE RANGE: 0 to {total_points} points (integers only)

STUDENT'S ANSWER:
{student_answer}

Provide your grading as:
Score: [integer from 0 to {total_points}]
Justification: [brief explanation based on rubric criteria]
\end{verbatim}
\end{qbox}

\clearpage
\section{Error Cases}
\label{app:error}
\subsection{Error Statistics} \Cref{tab:l1_errors_by_topic_styled,tab:l2_errors_by_topic_styled,tab:l3_errors_by_topic_styled} provide a quantitative breakdown of model errors across all three CFA MCQ levels. The tables categorize the total number of incorrect answers by topic for each model. Notably, while newer models significantly reduce overall errors, topics like Ethical and Professional Standards remain a persistent challenge for nearly all models on Levels I and II. For Level III, errors are more distributed, though topics like Performance Measurement and Derivatives and Risk Management are difficult for even the most advanced models.

\begin{table}[htbp]
\centering
\small
\caption{Breakdown of Chain of Thought errors on the Level I MCQ exam. Cell format: {Error Count (Error Rate \%)}. The error rate is calculated based on the total questions (n) per topic.}
\label{tab:l1_errors_by_topic_styled}
\begin{adjustbox}{max width=\textwidth}
\begin{tabular}{l|rrrrrrrrr}
\toprule
\bfseries Topic (n)
 & \multicolumn{1}{c}{\bfseries ChatGPT}
 & \multicolumn{1}{c}{\bfseries GPT-4}
 & \multicolumn{1}{c}{\bfseries GPT-4o}
 & \multicolumn{1}{c}{\bfseries GPT-5}
 & \multicolumn{1}{c}{\bfseries G2.5P}
 & \multicolumn{1}{c}{\bfseries G3P}
 & \multicolumn{1}{c}{\bfseries DeepSeek}
 & \multicolumn{1}{c}{\bfseries Grok 4}
 & \multicolumn{1}{c}{\bfseries Claude} \\
\midrule
Ethics (81)                      & 41 (51\%) & 24 (30\%) & 12 (15\%) & 4 (5\%)  & 7 (9\%)  & 3 (4\%)   & 19 (23\%) & 10 (12\%) & 12 (15\%) \\
Quantitative Methods (43)        & 8 (19\%)  & 5 (12\%)  & 0 (0\%)   & 0 (0\%)  & 1 (2\%)  & 0 (0\%)   & 0 (0\%)   & 0 (0\%)   & 0 (0\%)   \\
Economics (41)                   & 10 (24\%) & 4 (10\%)  & 4 (10\%)  & 3 (7\%)  & 3 (7\%)  & 1 (2\%)   & 2 (5\%)   & 2 (5\%)   & 2 (5\%)   \\
Financial Reporting (66)         & 25 (38\%) & 14 (21\%) & 7 (11\%)  & 1 (2\%)  & 2 (3\%)  & 3 (5\%)   & 4 (6\%)   & 2 (3\%)   & 2 (3\%)   \\
Corporate Issuers (41)           & 12 (29\%) & 4 (10\%)  & 4 (10\%)  & 1 (2\%)  & 1 (2\%)  & 2 (5\%)   & 4 (10\%)  & 1 (2\%)   & 3 (7\%)   \\
Equity Investments (67)          & 28 (42\%) & 11 (16\%) & 6 (9\%)   & 2 (3\%)  & 1 (1\%)  & 2 (3\%)   & 5 (7\%)   & 3 (4\%)   & 3 (4\%)   \\
Fixed Income (65)                & 20 (31\%) & 18 (28\%) & 8 (12\%)  & 3 (5\%)  & 1 (2\%)  & 0 (0\%)   & 2 (3\%)   & 1 (2\%)   & 4 (6\%)   \\
Derivatives (43)                 & 18 (42\%) & 9 (21\%)  & 5 (12\%)  & 0 (0\%)  & 1 (2\%)  & 2 (5\%)   & 3 (7\%)   & 0 (0\%)   & 0 (0\%)   \\
Alternative Investments (36)     & 9 (25\%)  & 6 (17\%)  & 4 (11\%)  & 2 (6\%)  & 1 (3\%)  & 1 (3\%)   & 2 (6\%)   & 1 (3\%)   & 0 (0\%)   \\
Portfolio Management (57)        & 18 (32\%) & 8 (14\%)  & 1 (2\%)   & 2 (4\%)  & 3 (5\%)  & 2 (4\%)   & 6 (11\%)  & 2 (4\%)   & 2 (4\%)   \\
\midrule
\textbf{Total (540)}             & \textbf{192 (36\%)} & \textbf{103 (19\%)} & \textbf{51 (9\%)} & \textbf{18 (3\%)} & \textbf{21 (4\%)} & \textbf{16 (3\%)} & \textbf{47 (9\%)} & \textbf{22 (4\%)} & \textbf{28 (5\%)} \\
\bottomrule
\end{tabular}
\end{adjustbox}
\vspace{-0.1in}
\end{table}

\begin{table}[htbp]
\centering
\small
\caption{A breakdown of Chain of Thought errors on the Level II MCQ exam. Cell format: {Error Count (Error Rate \%)}.}
\label{tab:l2_errors_by_topic_styled}
\begin{adjustbox}{max width=\textwidth}
\begin{tabular}{l|rrrrrrrrr}
\toprule
\bfseries Topic (n)
 & \multicolumn{1}{c}{\bfseries ChatGPT}
 & \multicolumn{1}{c}{\bfseries GPT-4}
 & \multicolumn{1}{c}{\bfseries GPT-4o}
 & \multicolumn{1}{c}{\bfseries GPT-5}
 & \multicolumn{1}{c}{\bfseries G2.5P}
 & \multicolumn{1}{c}{\bfseries G3P}
 & \multicolumn{1}{c}{\bfseries DeepSeek}
 & \multicolumn{1}{c}{\bfseries Grok 4}
 & \multicolumn{1}{c}{\bfseries Claude} \\
\midrule
Ethics (24)                      & 13 (54\%) & 6 (25\%)  & 8 (33\%)  & 4 (17\%) & 5 (21\%) & 5 (21\%)  & 11 (46\%) & 11 (46\%) & 5 (21\%) \\
Quantitative Methods (12)        & 2 (17\%)  & 0 (0\%)   & 1 (8\%)   & 0 (0\%)  & 0 (0\%)  & 0 (0\%)   & 0 (0\%)   & 0 (0\%)   & 0 (0\%)  \\
Economics (12)                   & 9 (75\%)  & 5 (42\%)  & 4 (33\%)  & 0 (0\%)  & 0 (0\%)  & 0 (0\%)   & 0 (0\%)   & 0 (0\%)   & 0 (0\%)  \\
Financial Reporting (24)         & 13 (54\%) & 6 (25\%)  & 7 (29\%)  & 3 (13\%) & 3 (13\%) & 2 (8\%)   & 4 (17\%)  & 2 (8\%)   & 3 (13\%) \\
Corporate Issuers (12)           & 6 (50\%)  & 2 (17\%)  & 1 (8\%)   & 1 (8\%)  & 1 (8\%)  & 1 (8\%)   & 1 (8\%)   & 0 (0\%)   & 2 (17\%) \\
Equity Investments (24)          & 7 (29\%)  & 2 (8\%)   & 1 (4\%)   & 0 (0\%)  & 0 (0\%)  & 2 (8\%)   & 0 (0\%)   & 0 (0\%)   & 0 (0\%)  \\
Fixed Income (24)                & 14 (58\%) & 8 (33\%)  & 9 (38\%)  & 2 (8\%)  & 2 (8\%)  & 2 (8\%)   & 1 (4\%)   & 3 (13\%)  & 2 (8\%)  \\
Derivatives (12)                 & 8 (67\%)  & 5 (42\%)  & 3 (25\%)  & 1 (8\%)  & 0 (0\%)  & 2 (17\%)  & 1 (8\%)   & 2 (17\%)  & 2 (17\%) \\
Alternative Investments (12)     & 6 (50\%)  & 5 (42\%)  & 3 (25\%)  & 0 (0\%)  & 0 (0\%)  & 0 (0\%)   & 0 (0\%)   & 1 (8\%)   & 0 (0\%)  \\
Portfolio Management (20)        & 6 (30\%)  & 8 (40\%)  & 4 (20\%)  & 1 (5\%)  & 0 (0\%)  & 0 (0\%)   & 4 (20\%)  & 2 (10\%)  & 1 (5\%)  \\
\midrule
\textbf{Total (176)}             & \textbf{91 (52\%)} & \textbf{53 (30\%)} & \textbf{46 (26\%)} & \textbf{13 (7\%)} & \textbf{13 (7\%)} & \textbf{14 (8\%)} & \textbf{25 (14\%)} & \textbf{24 (14\%)} & \textbf{18 (10\%)} \\
\bottomrule
\end{tabular}
\end{adjustbox}
\vspace{-0.1in}
\end{table}

\begin{table}[htbp]
\centering
\small
\caption{A breakdown of Chain of Thought errors on the Level III MCQ exam. Cell format: {Error Count (Error Rate \%)}. The error rate is calculated based on the total questions (n) per topic.}
\label{tab:l3_errors_by_topic_styled}
\begin{adjustbox}{max width=\textwidth}
\begin{tabular}{l|rrrrrrrrr}
\toprule
\bfseries Topic (n)
 & \multicolumn{1}{c}{\bfseries ChatGPT}
 & \multicolumn{1}{c}{\bfseries GPT-4}
 & \multicolumn{1}{c}{\bfseries GPT-4o}
 & \multicolumn{1}{c}{\bfseries GPT-5}
 & \multicolumn{1}{c}{\bfseries G2.5P}
 & \multicolumn{1}{c}{\bfseries G3P}
 & \multicolumn{1}{c}{\bfseries DeepSeek}
 & \multicolumn{1}{c}{\bfseries Grok 4}
 & \multicolumn{1}{c}{\bfseries Claude} \\
\midrule
Asset Allocation (12)            & 7 (58\%) & 6 (50\%) & 4 (33\%) & 3 (25\%) & 3 (25\%) & 3 (25\%) & 0 (0\%) & 3 (25\%) & 3 (25\%) \\
Portfolio Construction (24)      & 8 (33\%) & 7 (29\%) & 12 (50\%) & 4 (17\%) & 0 (0\%) & 6 (25\%) & 5 (21\%) & 7 (29\%) & 5 (21\%) \\
Performance Measurement (12)     & 9 (75\%) & 4 (33\%) & 5 (42\%) & 9 (75\%) & 5 (42\%) & 8 (67\%) & 7 (58\%) & 5 (42\%) & 7 (58\%) \\
Derivatives \& Risk (24)         & 12 (50\%) & 9 (38\%) & 10 (42\%) & 5 (21\%) & 2 (8\%) & 3 (13\%) & 2 (8\%) & 2 (8\%) & 7 (29\%) \\
Ethical Standards (12)           & 8 (67\%) & 7 (58\%) & 0 (0\%) & 4 (33\%) & 1 (8\%) & 0 (0\%) & 3 (25\%) & 4 (33\%) & 5 (42\%) \\
Portfolio Management (16)        & 10 (63\%) & 5 (31\%) & 7 (44\%) & 5 (31\%) & 4 (25\%) & 2 (13\%) & 2 (13\%) & 6 (38\%) & 4 (25\%) \\
Private Markets (16)             & 5 (31\%) & 3 (19\%) & 1 (6\%) & 0 (0\%) & 1 (6\%) & 0 (0\%) & 1 (6\%) & 1 (6\%) & 1 (6\%) \\
Private Wealth (16)              & 4 (25\%) & 4 (25\%) & 3 (19\%) & 3 (19\%) & 2 (13\%) & 4 (25\%) & 4 (25\%) & 1 (6\%) & 2 (13\%) \\
\midrule
\textbf{Total (132)}             & \textbf{63 (48\%)} & \textbf{45 (34\%)} & \textbf{41 (31\%)} & \textbf{33 (25\%)} & \textbf{18 (14\%)} & \textbf{26 (20\%)} & \textbf{24 (18\%)} & \textbf{29 (22\%)} & \textbf{34 (26\%)} \\
\bottomrule
\end{tabular}
\end{adjustbox}
\vspace{-0.1in}
\end{table}

\newpage
\subsection{Error Examples} 
\begin{figure}[H]
\makebox[\textwidth][c]{%
    \includegraphics[page=1,width=1.267\textwidth]%
    {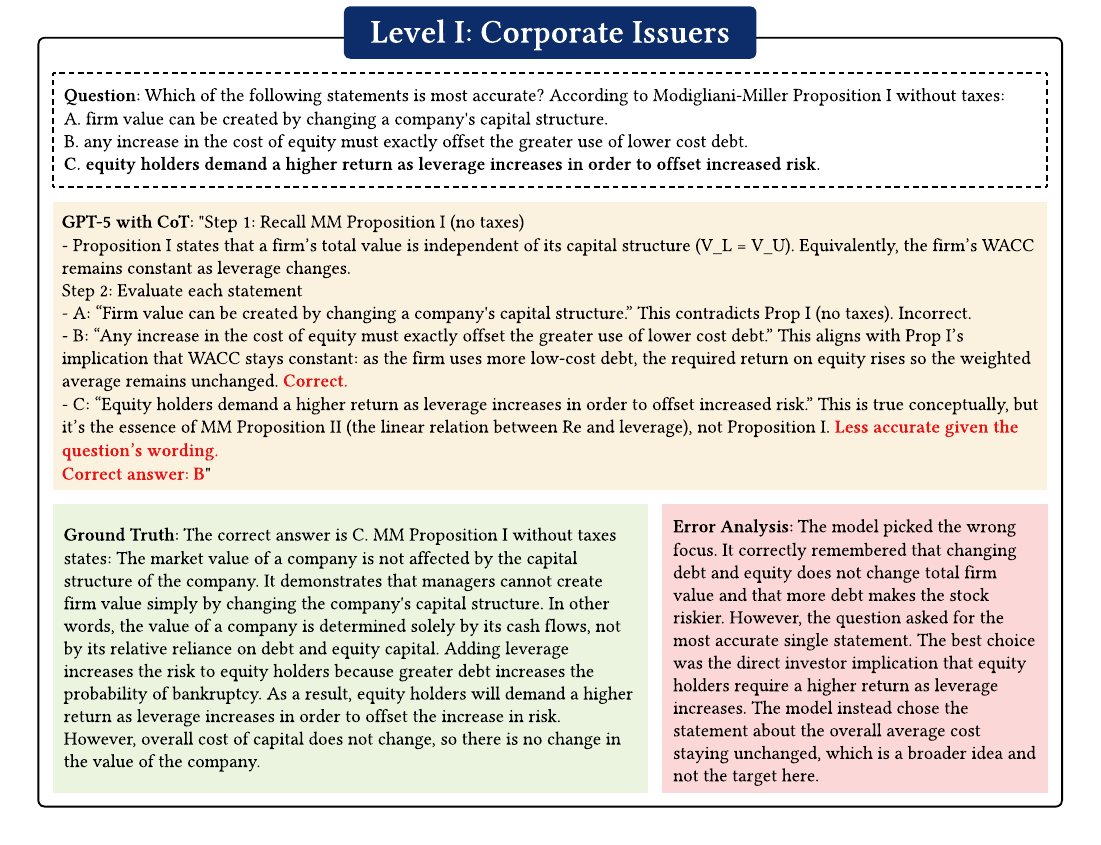}
}
\caption{Example of a Concept Misapplication error, where the model incorrectly selects between two related propositions.}
\label{fig:error_concept}
\vspace{-0.2in}
\end{figure}

\begin{figure}[H]
\centering
\makebox[\textwidth][c]{%
    \includegraphics[page=1,width=1.267\textwidth]%
        {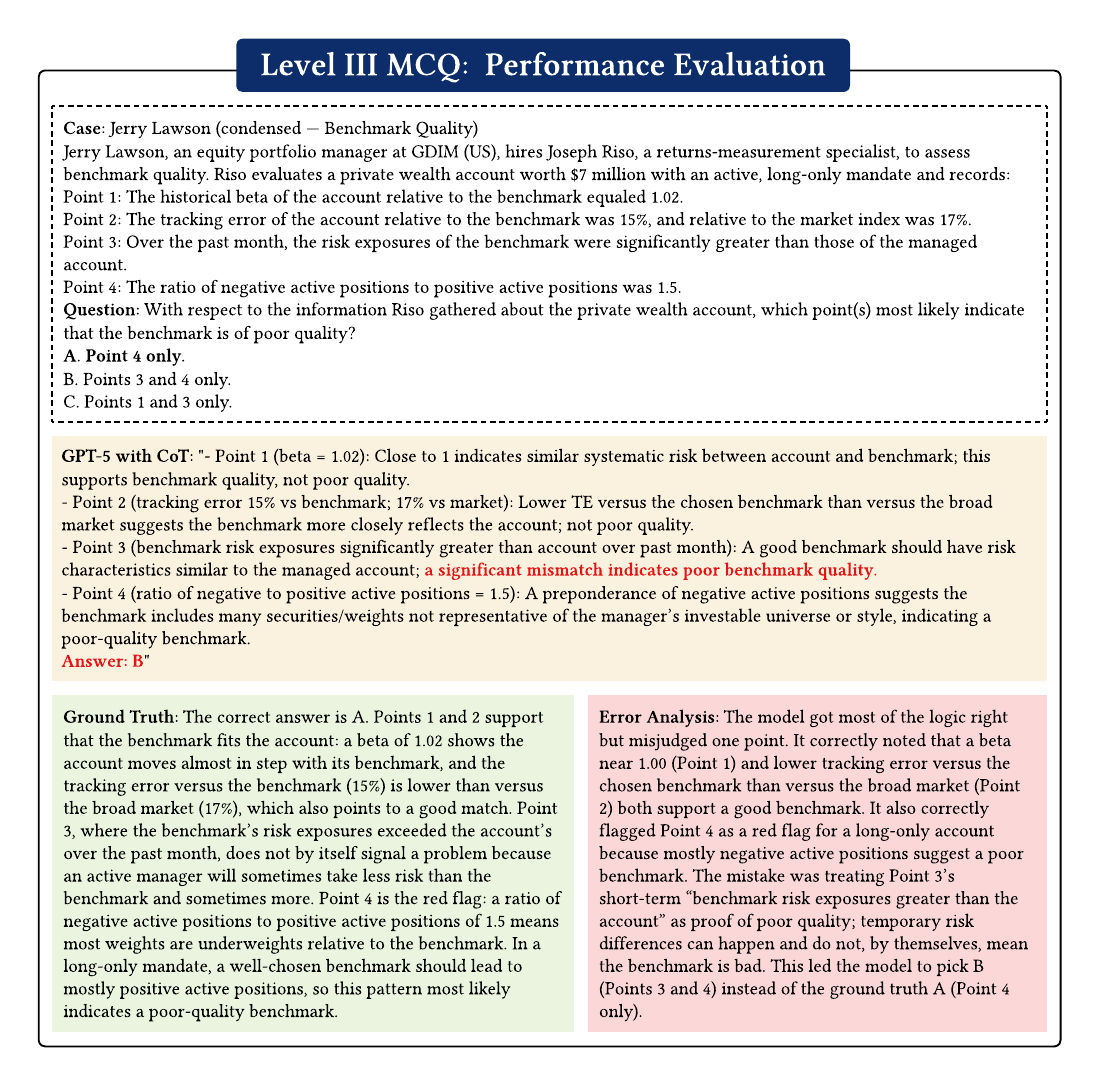}
}
 \caption{Example of a Misinterpretation of Evidence error, where the model incorrectly flags a normal portfolio activity as a sign of a poor benchmark.}

\label{fig:error_evidence}
\vspace{-0.2in}
\end{figure}

\Cref{fig:error_concept,fig:error_rule,fig:error_evidence,fig:error_oversimplification,fig:error_calculation} provide qualitative examples of common errors. Each figure presents a complete problem. This may include a case vignette (for Level II and III), a table, or a grading rubric (for CRQs). Each figure then shows the question (for MCQs, the correct answer choice is bolded), the model's incorrect response, the ground truth answer, model grading (for CRQs), and a brief error analysis. The following examples illustrate the specific types of logical flaws that persist in GPT-5.

\begin{figure}[H]
\centering
\makebox[\textwidth][c]{%
    \includegraphics[page=2,width=1.267\textwidth]%
        {images/Error_Examples/Error_Analysis_Figures.pdf}
}
\caption{Example of a Rule Application error, where the model misapplies ethical standards to a specific case vignette.}
\label{fig:error_rule}
\vspace{-0.2in}
\end{figure}

\begin{figure}[H]
\centering
\makebox[\textwidth][c]{%
    \includegraphics[page=2,width=1.267\textwidth]%
        {images/Error_Examples/Error_Analysis_Figures_level_iii.pdf}
}
\caption{Example of a Concept Oversimplification error, where the model provides a common but incorrect generalization instead of the nuanced answer.}
\label{fig:error_oversimplification}
\vspace{-0.2in}
\end{figure}

\begin{figure}[H]
\centering
\makebox[\textwidth][c]{%
    \includegraphics[page=3,width=1.267\textwidth]%
        {images/Error_Examples/Error_Analysis_Figures_level_iii.pdf}
}
\caption{Example of a Calculation Error, where the model uses incorrect base values for calculating the hurdle amount and profit, leading to an incorrect carried interest.}
\label{fig:error_calculation}
\vspace{-0.2in}
\end{figure}

\end{document}